\PassOptionsToPackage{numbers}{natbib}
\documentclass{article}

 \usepackage[preprint]{neurips_2025_ml4ps}

\usepackage{subcaption}
\usepackage[utf8]{inputenc} 
\usepackage[T1]{fontenc}    
\usepackage{hyperref}       
\usepackage{url}            
\usepackage{booktabs}       
\usepackage{amsfonts}       
\usepackage{nicefrac}       
\usepackage{microtype}      
\usepackage{xcolor}         
\usepackage{amsmath}
\usepackage{graphicx}
\title{Towards Universal Neural Operators through Multiphysics Pretraining}

\author{%
  Mikhail Masliaev \\
  ITMO University \\
  St. Petersburg, Russia, 197101\\
  \texttt{maslyaitis@gmail.com} \\
  \And
  Dmitry A. Gusarov \\
  ITMO University \\
  St. Petersburg, Russia, 197101\\
  \texttt{gusdmitr@itmo.ru} \\  
  \AND
  Ilya Markov \\
  ITMO University \\
  St. Petersburg, Russia, 197101\\
  \texttt{iomarkov@itmo.ru} \\
  \And
  Alexander Hvatov \\
  ITMO University \\
  St. Petersburg, Russia, 197101\\
  \texttt{alex\_hvatov@itmo.ru} \\
}

\begin{document}

\maketitle

\begin{abstract}
  Although neural operators are widely used in data-driven physical simulations, their training remains computationally expensive. Recent advances address this issue via downstream learning, where a model pretrained on simpler problems is fine-tuned on more complex ones. In this research, we investigate transformer-based neural operators, which have previously been applied only to specific problems, in a more general transfer learning setting. We evaluate their performance across diverse PDE problems, including extrapolation to unseen parameters, incorporation of new variables, and transfer from multi-equation datasets. Our results demonstrate that advanced neural operator architectures can effectively transfer knowledge across PDE problems.
\end{abstract}

\section{Introduction}

Contemporary science commonly uses partial differential equations (PDEs) and systems of partial differential equations to model spatio-temporal processes. For instance, reaction-diffusion equations describe, how mass moves and disperses within a fluid system, and gas and liquid dynamics are commonly modeled with variants of Euler/Navier-Stokes equations. While the analytical solutions are applicable in idealized problem statement, it is challenging to construct them in realistic scenarios. Thus, numerical simulation techniques, such as finite-element or spectral methods have been developed, yet in many cases such solutions can be computationally costly. However, in many cases, such as meteorological forecasting or multi-physics simulation in the engineering, the numerical solution of differential equations tends to be a computationally costly procedure.

With the development of scientific machine learning, greater emphasis in physics systems simulations is placed on data-driven methods. Physics-informed neural networks (PINN) \cite{raissi2019physics} extend the loss with PDE-based terms, so the training has an objective of matching network's output with governing PDE. However, PINNs require explicit formulations and guarantee accuracy only at training mesh nodes. Operator learning, realized through Deep Operator Networks (DeepONet) \cite{lu2021learning} and kernel-based neural operators (NO) \cite{kovachki2023neural}, offers a faster alternative to classical solvers, approximating mappings between functional spaces rather than dynamics at discrete nodes, providing discretization invariance and efficient inference.

A recent direction in NO research is the design of foundation models. Originating in NLP \cite{bommasani2021opportunities} or vision-language models \cite{awais2025foundation}, such models contain billions of parameters (e.g., GPT-4 exceeds one trillion \cite{achiam2023gpt}) and are pretrained on large-scale datasets. They can then be fine-tuned for downstream tasks at reduced cost. In this work, we aim to develop a standardized approach to applying large NO-based models for generalized dynamics and transfer learning across PDE problems. Thus, with the neural operator the model is pre-trained on a simplified problem statement, which is later transferred to another (typically, more complex) problem in the fine-tuning phase.

{\bf Our contributions are as follows:} the proposed method enables neural operator learning on diverse multi-physics datasets by introducing an adapter-based approach for simultaneous training on PDE-based problems with different sets of input functions. The results demonstrate that the transfer learning approach can significantly enhance model quality and reduce fine-tuning costs. 

\section{Related work}

Pretraining of neural operators has thus far mainly been case-specific, with limited generalization. The concept of transfer learning using DeepONet operator models are employed in conditional shift scenarios in study \cite{goswami2022deep}. Also, issues of solving transfer learning problems on multi-scale data with convolutional neural networks were discussed in \cite{subel2023explaining}.

Several foundational models for PDE systems have been proposed beyond classical NO approaches: a foundational model tranining framework for equations of fixed types (in the study, steady-state equations were considered) is proposed in the research \cite{subramanian2023towards}. Boundary-Embedded Neural Operators (BENO) \cite{wang2024beno} solve elliptic PDEs using graph neural networks, where boundary geometry is encoded via a transformer block into latent vectors guiding message passing. Other works leverage transformer-based architectures to encode PDE structure \cite{zhang2024deciphering}.

Transformer-based approaches proved to be capable of modeling complex interactions between selected token functions. POSEIDON, a hierarchical vision transformer with shifted windows, applicable to transfer knowledge across Euler/Navier–Stokes cases \cite{herde2024poseidon}. Codomain Attention Neural Operator (CoDA-NO) \cite{rahman2024pretraining}, designed for multiphysics PDE transfer learning, employs codomain attention with function space dot product.

Transformers have also been explored within neural operator learning \cite{boya2024physics}. In contrast, we deliberately avoid physics-informed approaches and focus on assessing the capacity of neural operators to learn generalized dynamics purely from data samples.

\section{Method}

An operator learning framework has been developed for data-driven modeling of dynamical systems, governed by parametric partial differential equations $L_{p} \mathbf{u}(t, \mathbf{x}) = f(t, \mathbf{x})$ on the bounded domain $\mathcal{D}$ in a mesh-agnostic (albeit with some limitations, as examined in \cite{fanaskov2023spectral}) approach. 


\paragraph{Neural operator learning:} In this research, we have focused on the kernel-integral neural operators, mainly Fourier Neural Operators. Neural Operators are designed to learn mappings between the input space $\mathcal{U}$, representing sets of input functions $\mathbf{a} = \{a_1, \; ... \; a_{n\_in}\}, \; a_i: \mathcal{D} \longrightarrow \mathbb{R}$ and the output functions $\mathbf{u} = \{u_1, \; ... \; u_{n\_out}\}, \; u_i: \mathcal{D} \longrightarrow \mathbb{R}$. While the outputs are typically fixed to the dependent variables of the PDE (or system of PDEs), the choice of input functions is guided by the equation structure and includes meshes, initial conditions, forcing terms, or equation coefficients. 

To respect the non-localities of the model, NO design adds integral kernel operators $(\mathcal{K}(v))(x)$ to the arguments ($A_t$, and $b_t$ for weights and biases) of activation functions $\sigma$. Here the $\kappa_{t} \in C(D_{t+1} \times D_{t} | \theta_{k, t}); \, \theta_{k, t} \in \mathbb{R}^{n_{v_{t+1}} \times n_{v_{t}}}$ is the kernel function, parameterized (with $\theta_{k, t}$) by the method of choice (FNO, GNO, etc.), and $v_t(y)$ - hidden representation of input functions, obtained from the $t$-th layer. $D_{t}$ is the hidden dimensionality, which is linked to the number of modes in FNO. Thus, the parameters of NO main part include weights, biases and kernel parameters $\theta_{\mathcal{F}} = \{A_{t}, b_{t}, \theta_{k, t}:  t = 1, \; ... \;, n_{\text{layers}}\}$.

\begin{equation}
\label{eq:fno_layer}
    \mathcal{F}_{t}(x) = \sigma \left( A_t v_t(x) + \int_{D_{i}} \kappa_{t}(x,y) v_{t}(y) dy + b_t(x)\right), \; \forall x \in D_{t}, \; t = 1, \; ... \;, n_{\text{layers}}\; 
\end{equation}

The architecture of layers sequence in NO goes as follows: the inputs are transformed to their higher-dimensional hidden representation by lifting layers (typically, a feed-forward neural network) $\mathcal{L}: \mathcal{L}(\mathbf{a}) = \sigma \left( A_{\mathcal{L}} \mathbf{a} + b_{\mathcal{L}} \right)$, where parameters of the lifting include weights and biases $\theta_\mathcal{L} = \{A_{\mathcal{L}}, b_{\mathcal{L}}\}$. The hidden representations are sequentially mapped with the integral-operator blocks \eqref{eq:fno_layer}. In contrast, the output of the last block is projected to the space of outputs by the point-wise function $\mathcal{P}$ with parameters $\theta_\mathcal{L} = \{A_{\mathcal{P}}, b_{\mathcal{P}}\}$. The operator approximation takes form of model $\mathcal{G}_{\theta}$: $\tilde{\mathbf{u}}(\mathbf{x}) = \mathcal{G}_{\theta} (\mathbf{a}) = \mathcal{P}  \; \circ \; \mathcal{F} \; \circ \mathcal{L}(\mathbf{a}) = \; \mathcal{P} \; \circ \; \mathcal{F}_{n\_layers}  \; \circ \; ... \; \circ \; \mathcal{F}_{1} \; \circ \; \mathcal{L}(\mathbf{a})$.








\paragraph{Improving the generalization ability of neural operators:} In this research, we employed two types of NO modifications: state-space models and transformer-based models. In the first approach, inserting a Mamba-SSM module \cite{gu2023mamba} $\mathcal{M}_{\phi}$ after the lifting map $\mathcal{L}$ allows the model to encode long-range temporal and spatial dependencies directly in the hidden representation. For lifted features $v_{0}(x)=\mathcal{L}(\mathbf{a})(x)$, the Mamba module computes

\begin{equation}
\widetilde{v}_{0}(x,t)=(\mathcal{M}_{\phi} v_{0})(x,t)=\sum_{\tau \le t} K_{\tau}\,v_{0}(x,t-\tau),    
\end{equation}

with learnable convolution kernels $K_{\tau}$ defining the causal recurrence. This step acts as a latent preconditioner: embeddings are aligned with dominant dynamical motifs (transport, diffusion, oscillation) common across PDEs, so that when passed into the Fourier integral layers, the effective operator acts on inputs of reduced variability and lower spectral rank. Consequently, the composition $\mathcal{F}_{t}\circ \mathcal{M}_{\phi}$ yields more stable training and improves efficiency in transferring pre-trained representations to new PDEs during fine-tuning.


The next approach, examined in the study, involved the attention method \& transformer-based blocks. 
The introduction of Perceiver \cite{jaegle2021perceiver} enabled the encoding of information with a smaller number of latent feature arrays, internal to operator blocks, thereby operating with more abstract feature arrays and maintaining a limited number of parameters.
As the operators we employ, we use blocks based on the Perceiver IO \cite{jaegle2021perceiverio}, where the mapping is performed with a symmetrical cross-attention mechanism for outputs, which mirrors the cross-attention block for constructing representations of latent arrays from the input process. While the previously used self-attention blocks can discover dependencies between hidden features, obtained from lifting or previous layers, Perceivers are able to constructs additional latent process representation.

The latent variables and input state are combined first with the cross-attention block, where keys and values are obtained from FNO-based mapping from the inputs $K_1 = FNO_{K_1}(X)$, $V_1 = FNO_{V_1}(X)$, and latent variables are taken as queries $Q_1 = L$. The cross-attention block is followed by self-attention between latent representation. The output of the block is constructed with the cross-attention, matching the queries from the inputs with the keys and values, taken from the transformed latent representations.

Commonly used self-attention mechanism involves similarity function $\text{sim}(q_m, k_j)$ between given sets of finite-dimensional vectors of queries $\{\mathbf{q}_{i}\}, \; i = 0 , \; ... \;, N_q$ and keys $\{\mathbf{k}_{i}\}, \; i = 0 , \; ... \;, N_k$, which is used to obtain the output from the $m$-th query with the set of value vectors $\{\mathbf{v}_{i}\}, \; i = 0 , \; ... \;, N_v$ with the relation. The similarity is commonly obtained using \textit{Softmax} function of the dot products between corresponding queries and keys. Codomain attention mechanisms, introduced in \cite{rahman2024pretraining}, are advantageous to the conventional transformers in the neural-operator based problems: the dot product detecting similarity not between samples, but between features, mapped with neural operators.    




\paragraph{Pre-training and fine-tuning:} One of the benefits of using lifting-operator-projection architecture is the simplicity of decoupling adapters from the main model, streamlining model storage and extension for novel fine-tuning problems. The \texttt{lift} and \texttt{proj} blocks are considered as the adapters, representing the mappings, associated with the problem-specific part of dynamics: they are introduced to contain different cardinality input sets, projecting into the fixed number of hidden features and contain small number of parameters to represent limited part of the total model variance, as it is common in the adapter design for large language models \cite{hu2022lora}.

In the pre-training phase the entire parameters set $(\theta_{\mathcal{P}_{1}}, \; ... \;, \theta_{\mathcal{P}_{N}}, \theta_{\mathcal{F}}, \theta_{\mathcal{L}_{1}}, \; ... \;, \theta_{\mathcal{L}_{N}})$ is subject to optimization. By problems $1$ to $N$, we present separate physical processes, demanding different (but, probably, overlapping) sets of input functions. Previously, such problems were solved with liftings with extensive inputs. For example, in training on the set of steady-state problems, coefficients before terms with all spatial derivatives, which occur in the training set, are used as inputs. 
In the fine-tuning stage we fix the parameters $\theta_{\mathcal{F}}$ both to highlight the generalizing properties of the operator and to reduce training costs: only the new adapter parameters $(\theta_{\mathcal{P}_{ft}}, \theta_{\mathcal{L}_{ft}})$ are trained.

\begin{figure}
\centering
\includegraphics[width=0.99\textwidth]{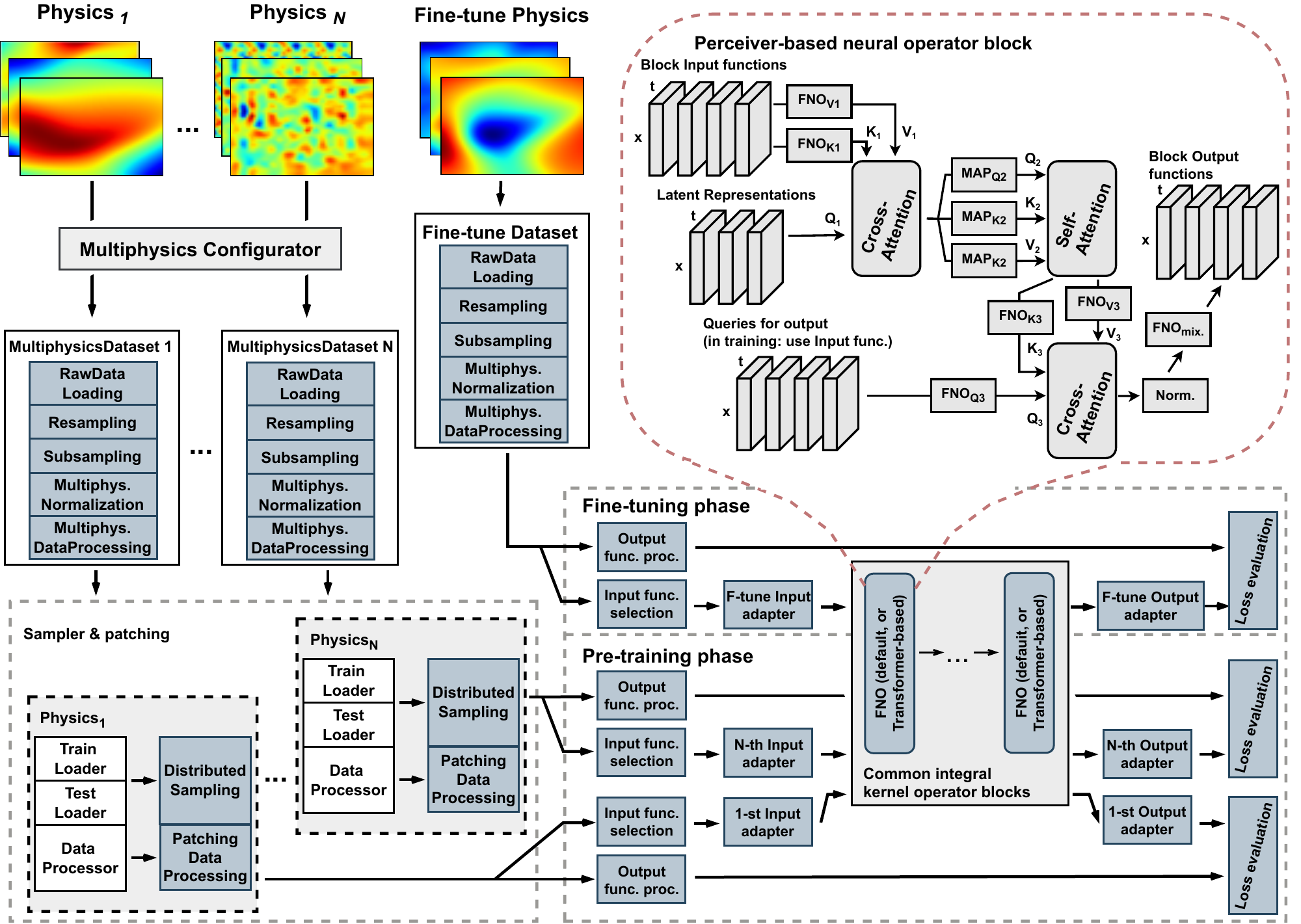}
\caption{Scheme of the neural operator pre-training and fine-tuning stages. In this scheme, FNO blocks denote arbitrary kernel integral operators, including the transformer-based architectures. The separate physics $1$ to $N$ may vary from the different manifestations of the same system to multiple different physics (but with the same problem dimensionality) in the dataset collection.}
\label{fig:pre-train_fine-tune_sketch}
\end{figure}

\section{Experiments}

We aim on validating our approach on three distinct types of problems without modifications in the modeling approach. The first scenario involves cases, when the pre-training and fine-tuning processes are governed by the equations with same input functions, but with different parameters. In addition to parametric variations within a single physical law, we further extend our investigation to cross-domain transfer learning between classes of differential equations representing distinct physical phenomena. 
We have selected datasets to represent diverse physical phenomena, including advective transport processes, nonlinear wave dynamics governed by Burgers equation, reaction-diffusion systems exhibiting pattern formation, and other fundamental physical processes described by partial differential equations.


\begin{equation}
\mathrm{NMAE}(\theta)
=
\frac{1}{|\mathcal{D}_{\mathrm{RD}}^{\mathrm{test}}|}
\sum_{(\mathbf{a},u)\in\mathcal{D}_{\mathrm{RD}}^{\mathrm{test}}}
\frac{\left\| \mathcal{G}_{\theta}(\mathbf{a}) - u \right\|_{1,G}}
     {\max_{G} u - \min_{G} u + \varepsilon}
\label{eq:nmae}
\end{equation}

In this comparison, we employ developed post-lifting (PL) MambaFNO models, post-lifting (PL) LocalAttnFNO, and Perceiver IO-based NO, as well as Swin-v2 transformers and CodaNO models. As the baseline, we employed the default FNO. We used the range-normalized mean absolute error (NMAE) \eqref{eq:nmae} as a quality metric. The code and experiments are available in the repository \url{https://anonymous.4open.science/r/multiphysics_neurop-F385/}.

\paragraph{Out-of-sample parameter values scenario}

First, we conducted several experiments on cases where the pretraining equations and fine-tuning ones differed only in the coefficient values. The experiments were conducted using Burgers' equation , the Gray-Scott model of the reaction-diffusion process , and the Navier-Stokes equations for an incompressible flow. 
The results of the comparison are presented in Tab.~\ref{tab:oos_exp}.

\begin{table}[ht!]
\centering
\caption{Average metric values for out-of-sample parameter values across all conducted experiments. Methods were compared with training from scratch on the examined datasets scenario and using the pre-trained model to fine-tune on the new dynamics.}
\resizebox{0.75\columnwidth}{!}{
\begin{tabular}{lcccc}
\toprule
\textbf{Model} & \textbf{MSE} & \textbf{NMAE (\%)} & \textbf{Avg. epoch (s)} & \textbf{Param.} \\
\midrule
Mamba FNO (pretr.)    & $1.009 \times 10^{-7}$ & 0.0120 & 21.91 & $\approx 10^7$ \\
Mamba FNO (scratch)   & $1.193 \times 10^{-7}$ & 0.0213          & 40.14 & $\approx 10^7$ \\
Perc. (pretr.)    & $1.425 \times 10^{-7}$ & 0.0169  & \textbf{3.21} & $\approx 10^8$ \\
Perc. (scratch)   & $1.981\times 10^{-7}$  & 0.0219  & 204.73 & $\approx 10^8$ \\
FNO (scratch)     & $1.774\times 10^{-7}$  & 0.0204  & 7.44 & $\approx 10^6$ \\
Swin-v2 (p.+s.)  & $4.391 \times 10^{-8}$ & \textbf{0.0092}  & 101.3 & $\approx 10^9$ \\
CoDA-NO (pretr.)  & $2.881 \times 10^{-7}$ & 0.0343  & 62.91 & $\approx 10^8$ \\
CoDA-NO (scratch) & $4.912 \times 10^{-7}$ & 0.0712  & 63.29 & $\approx 10^8$ \\
\bottomrule
\label{tab:oos_exp}
\end{tabular}
}
\end{table}


\paragraph{Input function set extension scenario \& General multi-physics learning}

To assess the applicability of the adapter-based approach, several experiments were conducted on scenarios where the equations were extended with additional terms. Here, for fine-tuning, we added convection to the heat equation and extended reaction-diffusion equations with advection. 

In the final stage, we evaluated the capabilities of the developed methods to transfer knowledge from the dynamics of advection and Burgers' equation to reaction–diffusion, based on the PDEBench dataset \cite{takamoto2022pdebench}. The combined results of the experiments are presented in Tab.~\ref{tab:full_exp}. The significant speedup achieved with the pre-trained models can be attributed to the optimization of just a subset of parameters, whereas "from scratch" involved a full parameter search.


\begin{table}[ht!]
\centering
\caption{Results of experiments with Heat \& Reaction–Diffusion equation extension, and multi-physics pre-training with fine-tuning on different dynamics.}
\resizebox{0.65\columnwidth}{!}{
\begin{tabular}{lccc}
\toprule
\textbf{Model} & \textbf{MSE} & \textbf{NMAE (\%)} & \textbf{Avg. epoch (s)} \\
\midrule
Mamba FNO (pretr.)    & $3.91 \times 10^{-6}$ & \textbf{0.0041} & 131.2 \\
Mamba FNO (scratch)   & $4.291 \times 10^{-6}$ & 0.0054 & 261.1  \\
Perc. (pretr.)    & $4.107 \times 10^{-6}$ & 0.0051  & \textbf{20.4} \\
Perc. (scratch)   & $6.315 \times 10^{-6}$  & 0.0074  & 804.0 \\
FNO (scratch)     & $7.286 \times 10^{-6}$  & 0.0121  & 41.3 \\
Swin-v2 (p.+s.)  & $6.276 \times 10^{-6}$ & 0.009  & 301.1 \\
CoDA-NO (pretr.)  & $1.043 \times 10^{-5}$ & 0.013  & 185.1 \\
CoDA-NO (scratch) & $1.239 \times 10^{-5}$ & 0.018  & 181.9 \\
\bottomrule
\label{tab:full_exp}
\end{tabular}
}
\end{table}

\section{Conclusion}

In this research, we have examined the performance of NO architectures, based on transformers and structured state space models, on a set of transfer learning problems involving changes to the parameters of equations, the inclusion of additional physics, and, finally, pre-training on multiphysics datasets. In contrast to the default NO approach, developed models have greater generalizing power. Use of parameter sets and problem-specific adapters enables the more effortless transfer of knowledge, reducing the cost of obtaining a decent model, even for novel PDE-based problems. 

This research was conducted primarily as the first stage of work towards a foundational model, pre-trained on vast and heterogeneous multiphysics dataset collections. The follow-up work shall be directed towards improving model performance and training, implementing data augmentation tools (both generic and PDE-specific, e.g., based on Lie symmetries), and further examining neural operator generalizations.


{
\small
\bibliographystyle{nips}
\bibliography{references.bib}
}

\end{document}